\documentclass{article}

\usepackage{arxiv}

\usepackage[utf8]{inputenc} 
\usepackage[T1]{fontenc}    
\usepackage{hyperref}       
\usepackage{url}            
\usepackage{booktabs}       
\usepackage{amsfonts}       
\usepackage{nicefrac}       
\usepackage{microtype}      
\usepackage{lipsum}
\usepackage{graphicx}
\usepackage{float}
\graphicspath{ {./Figures/} }

\newcommand{\beginsupplement}{%
        \setcounter{table}{0}
        \renewcommand{\thetable}{S\arabic{table}}%
        \setcounter{figure}{0}
        \renewcommand{\thefigure}{S\arabic{figure}}%
     }

\title{Translating synthetic natural language to database queries: a polyglot deep learning framework}

\author{
 Adrián Bazaga \\
  Department of Genetics\\
  University of Cambridge\\
  Cambridge, United Kingdom \\ \\
  and\\ \\
  STORM Therapeutics Ltd\\
  Cambridge, United Kingdom\\
  \texttt{ar989@cam.ac.uk} \\
   \And
 Nupur Gunwant \\
  Department of Genetics\\
  University of Cambridge\\
  Cambridge, United Kingdom \\
  \And
 Gos Micklem \\
  Department of Genetics\\
  University of Cambridge\\
  Cambridge, United Kingdom \\
  \texttt{gm263@cam.ac.uk} \\
}

\begin{document}
\maketitle
\begin{abstract}
The number of databases as well as their size and complexity is increasing. This creates a barrier to use especially for non-experts, who have to come to grips with the nature of the data, the way it has been represented in the database, and the specific query languages or user interfaces by which data are accessed. These difficulties worsen in research settings, where it is common to work with many different databases. One approach to improving this situation is to allow users to pose their queries in natural language. 

In this work we describe a machine learning framework, Polyglotter, that in a general way supports the mapping of natural language searches to database queries. Importantly, it does not require the creation of manually annotated data for training and therefore can be applied easily to multiple domains. The framework is polyglot in the sense that it supports multiple different database engines that are accessed with a variety of query languages, including SQL and Cypher. Furthermore Polyglotter also supports multi-class queries.
 
Our results indicate that our framework performs well on both synthetic and real databases, and may provide opportunities for database maintainers to improve accessibility to their resources.
\end{abstract}


\section{Introduction}

Natural language querying (NLQ) is a known problem in information retrieval \cite{affolter_comparative_2019,dar_frameworks_2019}. It allows questions to be formed without knowledge of database-specific logical languages such as SQL or Cypher. In principle this can ease data access for non-expert users. To this end, several approaches to build NLQ systems have been proposed. Recent surveys \cite{affolter_comparative_2019,reshma_review_2017}, segmented them into five approaches: keyword-based, pattern-based, syntax-based, grammar-based, and, more recently, connectionist-based.

In keyword-based systems, the approach has two stages, a first stage where keywords present in an input query are extracted, and a second stage where keywords are matched against metadata available in the underlying database. For instance, \cite{blunschi_soda_2012} described Search over DAta Warehouse (SODA), which generates SQL queries from natural language (NL) queries over a business-related database. The system processes an input NL query through a series of steps. First, the keywords in the query are matched against all the possible entries in the database metadata. Second, by means of a heuristic, each result is scored and the process continues with the top N results. A third step identifies the tables used by each result and their relationships before being passed to the fourth step, which is responsible for finding, from the original query, the needed filters over the tables and columns. Last, the gathered information from the previous steps is combined to generate a SQL query that takes into account possible join patterns by looking at foreign keys and inheritance patterns in the schema. A drawback of this approach is it requires hand-crafting the patterns that are used to translate from a keyword-based input to a SQL query for the specific modelling of the target database, which becomes a bottleneck when trying to use it across a variety of databases.

In pattern-based systems, the capabilities of keyword-based approaches are extended by including NL patterns for processing queries, alleviating issues such as aggregation operations. For instance, \cite{shah_nlkbidb_2013} described NLKBIDB, a NL to SQL query interface that uses NL patterns to fix syntactically incorrect queries, and a keyword-based approach to obtain the corresponding facts from the schema before carrying out the conversion. The system uses lexical analysis to tokenize the NL query and syntax analysis to parse the lexicons. If the input query is syntactically valid, then the lexicons are analyzed by a semantic analyzer using a domain ontology before generating a SQL query. Invalid queries are converted into SQL by applying a set of rules. An issue that arises from this approach is the need for a knowledge expert to refine the rules used to convert syntactically invalid queries into SQL sentences, as well as the need to use hand-crafted natural language rules for generating the SQL queries for syntactically valid queries. This makes it difficult to adapt for new use cases.

Syntax-based and grammar-based systems \cite{franconi_panto_2007} share similar methodologies. In both approaches, the NL query is parsed using linguistic rules to produce a syntax tree. Then, for syntax-based systems, the concepts in the tree are mapped to a query in the target query language (e.g. SQL). This has a variety of issues. First, a given query may have different parse trees, which after mapping may produce different queries. Another issue is deciding which concepts from the query to map and which ones not. To alleviate this, grammar-based systems exploit domain knowledge captured by the grammar with the aim of reducing ambiguity when mapping queries. However, this method requires knowledge of the domain to build an effective parsing grammar, making it hard to adapt to new domains.

In contrast to approaches treating items in a language as symbols (symbolic approaches), relying on theoretical foundations in linguistics, connectionist approaches \cite{sinha_connectionist_1994}, also known as computational intelligence-based approaches, try to learn statistical patterns in the data and/or distributed representations of it, allowing for a richer linguistic variability. These methods can be divided into traditional machine learning and more recent deep learning techniques. The latter have become widely used in natural language processing (NLP) tasks, achieving state-of-the-art results \cite{deng_deep_2018,young_recent_2018}. The main difference between these methods and traditional machine learning is that their objective is to learn a distributed representation (in the form of real-valued vectors) of the data, without the need for the feature engineering stage that earlier approaches required.

Deep learning methodologies cover key NLP applications \cite{huang_deep_2019}, including part-of-speech tagging \cite{dichev_deep_2016}, named entity recognition \cite{habibi_deep_2017} and machine translation \cite{bahdanau_neural_2016}. In such systems the problem usually has been posed as an end-to-end neural semantic parsing problem \cite{dong_language_2016}. A specific case is the supervised sequence-to-sequence machine translation \cite{sutskever_sequence_2014} problem, in which the algorithm is trained to construct a representation encoding the input sentences in one domain (e.g. natural language) and decode them into output sentences in a different domain (e.g. a database specific language, such as SQL). 

For instance, the Seq2SQL system \cite{zhong_seq2sql_2017} is a seq2seq-based deep neural network for translating NL questions to their corresponding SQL queries, using a reinforcement learning policy to learn the conditions of the SQL query. The output space of the softmax function that predicts each token in the SQL query is limited by leveraging a priori knowledge of the database schema, thus reducing the number of invalid queries. The model is evaluated by running the predicted SQL query on the database and comparing the result obtained against the ground truth. The authors generated a dataset to train and evaluate their model, named WikiSQL, comprising 80,654 manually annotated example questions, and their equivalent SQL queries and results across 24,241 tables from Wikipedia. Despite achieving a highly accurate model, the system is intended to work on a single table at a time, with the system already knowing on which table it has to run the query as well as the schema describing the table. With a similar approach, \cite{xu_sqlnet_2017} describes SQLNet, which employs a sketch-based scheme containing a dependency graph, and a column attention mechanism to synthesize the query from the sketch. In a similar fashion to Seq2SQL, the SQLNet handles translation of NL queries to SQL over a single database table, taking a pair of the input table schema and the NL query. Thus this approach only provided single table queries based on SQL.

\cite{yin_neural_2016} described Neural Enquirer, a neural network architecture for predicting a ranked list of possible answers to a NL question, $Q$, over a SQL database table, $T$. The system embeds both $Q$ and $T$ into a distributed representation before passing it through a pipeline of executors that derive a probability distribution over the table entries as a ranked list of possible answers to the original NL query. In order to evaluate the system, the authors present an approach to generate synthetic question-table-answer triples on which the models are trained. In this generated dataset, each query involves a single table randomly sampled from an Olympic Games database. The natural language for the query is generated using NL templates to generate 4 different types of queries. The system reports an overall accuracy of 99.9\% on a synthetic dataset containing 100,000 query-table-answer triples. Like Seq2SQL and SQLNet, this system does not consider searching over multiple tables with a single query, and similarly is engineered to work only with SQL databases.

In an effort to alleviate the requirement to have a manually annotated dataset to train a NL to database (DB) interface, \cite{weir_dbpal_2019} developed DBPal, a deep neural network for learning natural language interfaces to SQL databases that synthesizes a large collection of pairs of NL queries and their corresponding SQL statements from a given database schema to train the model. The authors employ data augmentation by means of paraphrasing to make the model more robust to linguistic variation, and devise a manually crafted dataset of ~300 queries to assess the robustness of the model, where the same NL query is written with multiple linguistic variants.

In this paper, we propose a polyglot NL-to-query system, Polyglotter, to address some of the limitations in the approaches described above. Instead of being limited to queries across a single table, we support questions over multiple tables in a single query. We create synthetic data for training from the data models of each of the supported database engines. These synthetic datasets are then used to train a sequence-to-sequence-based algorithm. In order to allow Polyglotter to handle queries for a variety of database engines, we introduce an abstract representation for user queries in the form of a graph. This contains the attributes, classes and constraints, and acts as an intermediate representation that facilitates support for multiple database languages, e.g. SQL or Cypher. This makes it easier to re-use our system in new contexts. This graph can be decomposed into pairs (a class with any one of its attributes) and triples (a class with any one of its attributes and a constraint on that attribute) that forms a convenient target for machine learning. Note that a constraint is made up of an operator that makes a comparison and a value. The trained system is able, given a user query in natural language, to provide multiple suggestions in the specified database-specific logical language. We assess the performance of our system in three widely used database back-ends: MySQL, Neo4j and web service-based data storage, loaded with real and synthetic data from multiple domains: biology, business and product line management. With the above developments we hope to provide a generally useful system (1) that provides a simple and data-less way to build natural language interfaces supporting multi-table queries, (2) which supports multiple database engines and (3) that can be extended to support other database systems.

\section{Synthesizing training data using random query generation}

\begin{figure}[H]
  \centering
  \includegraphics[keepaspectratio,scale=0.6]{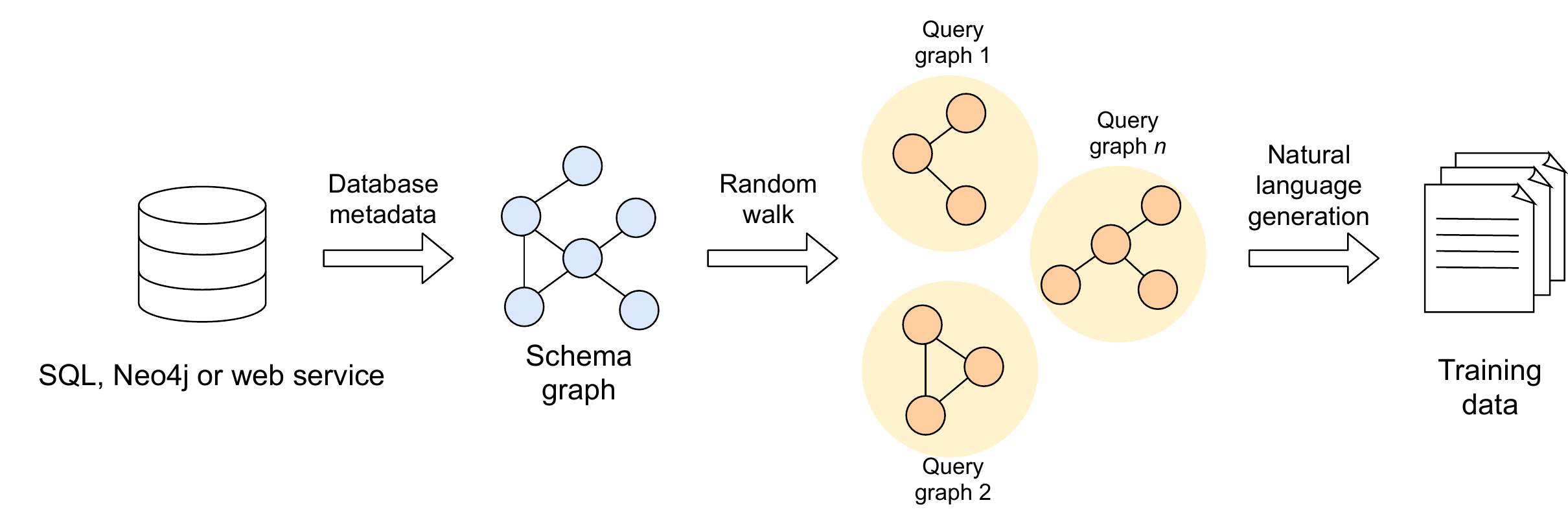}
  \caption{Overview of the random query generation method used by Polyglotter to obtain training datasets for the sequence-to-sequence models. Classes and attributes are added through a random walk, and constraint operators and values are also selected randomly.}
  \label{fig:Figure1}
\end{figure}

To generate the training dataset for a specific database backend, we derive a schema graph abstraction that represents the entities existing in the database, as well as the relations between them. Then we perform random walk-based transversals on it, as described below, to emit multiple subgraphs describing different queries over the database. These are further transformed to synthetic natural language to build the training set. Figure \ref{fig:Figure1} provides an overview of the steps followed by the query generation procedure.

We start by gathering the database metadata in order to collect four elements: the different entities existing in the database, the number of instances within them, their relationships and the attributes that describe each of them. To this extent, our implementation already provides an extensible set of metadata gatherers for several popular database engines: MySQL, Neo4j and web services / APIs. In the case of web services-based databases, we specifically provide an example implementation for HumanMine \cite{kalderimis_intermine_2014}, an open-source biological data warehouse for Homo sapiens data. 

In the schema graph, each node represents a class (e.g. a table in the case of MySQL), and nodes are connected if there is a corresponding class-class relationship (e.g. foreign keys in the case of MySQL). Also, each of the nodes is enriched with the attributes that it contains (e.g. the columns in the case of a node corresponding to a MySQL table). 

Once the database schema graph has been obtained, the query generation process performs random walks over it each of which selects a contiguous subset of the graph, which can be represented as a query. The random walk procedure is parametric, allowing the system maintainer to tune the complexity of the generated queries as described below. The number of query graphs to produce is controlled by the parameter $N$.

The starting node for a random walk is chosen randomly following a uniform distribution over classes. Queries can be enlarged by selecting which attributes of the class to report and also by applying constraints to the values that these attributes may take, as well as by adding further classes. The attribute and constraint choice probabilities ($attributeChoiceProbability$ and $constraintChoiceProbability$ within the code) control the likelihood of adding an attribute or constraint to the query for the current class. The constraint logic is picked randomly by rolling a dice over a distribution of possible logical operators. Similarly, the graph traversal probability parameter determines the chance of adding another class to the query. This is done by considering those directly connected nodes that have not already been added to the query. The $cap\_classes$ parameter sets a limit on the number of classes allowed in each query.

In our experiments we used the following values for the parameters: a uniformly random starting node selection, multi-class queries, 0.25 and 0.05 for the attribute and constraint choice probabilities respectively and 0.5 for the graph traversal probability. We limited the maximum number of classes in each query by taking into account the size of each database schema. For Neo4j, MySQL and HumanMine with schemas comprising a total of 5, 8 and 170 classes respectively, we capped the number of classes possible in a query to a maximum of 3, 4, or 5 respectively. We found these values provided a reasonable diversity of queries that looked realistic to a human.

Finally, once $N$ query graphs have been generated, each of them is translated into an equivalent query in English, and also into the target pairs and triples that the model will need to predict. Attributes from classes in the query, are modelled as $Class-Attribute$ pairs. Similarly, constraints over an attribute in a class are modelled as $Class-Attribute-Constraint$ triples. Note that the Constraint is itself a pair composed of Constraint logic and Constraint Value. The pairs and triples for a given target query are separated by the special token “$;$”. 

In simple terms, the translation to synthetic natural language is carried out by going over each of the elements within a query graph and expressing it in English. Several approaches are taken to increase the diversity of this generated English: the system randomly chooses one of six different styles of sentence, where each variation alters the order in which the different types of element (class, attribute, constraint) are dealt with. For example, one variation iterates over the classes, and with each describes the corresponding attributes and constraints; another begins with the attribute(s), followed by all the class(es) and finally all the constraint(s); a third one specifies first the constraints, then the attributes and then the classes. Further variability is generated by starting the conversion at different locations in the graph, and by picking synonyms of verbs and constraint operators randomly. In this way, we try to capture some of the diversity of expression that an English speaker would use (See Supplementary Table 1). Finally, the generated sentences are stored paired with their attributes, classes and constraints using a valid format for OpenNMT \cite{klein-etal-2017-opennmt,klein-etal-2018-opennmt,klein-etal-2020-opennmt}.

\section{Development of a neural sequence-to-sequence model with an attention mechanism}

We constructed a deep learning-based sequence-to-sequence model to predict the set of attributes, classes and constraints from a given user query, in the form of the pairs and triples as defined above. In this work we utilize the recently proposed Transformer architecture \cite{vaswani_attention_2017} to train a model with such capabilities. Apart from overall higher performance in this kind of task, one of the main advantages of this model, in contrast to the older Recursive Neural Network (RNN)-based models, is the highly parallelizable training phase, making training on large corpora more efficient.

The overall objective of a sequence-to-sequence model is to learn a latent representation (embedding) of an input language, in order to be capable of rebuilding the equivalent sentence in a different language or vocabulary from the learned embedding. The part of the model responsible for encoding a sentence from the input language is called the encoder, whereas the part that transforms the embedding to a sentence in the target language is called the decoder. The model uses the embedding to represent the semantic meaning of each word, as well as learning the most relevant elements in the input to be looked at as each word is examined in turn by using a self-attention mechanism.

The Transformer model improves upon the self-attention mechanism by introducing multi-headed attention, so learning a set of attention matrices instead of one. This provides two main benefits: first, it extends the ability of the model to focus on different positions, by helping to prevent the attention being dominated by the word translated at each timestep. Second, each of the attention sets is initialized with different random values, providing different representation subspaces when used during the projection of the input embeddings. Through an ablation study (Supplementary Figure 1), we found the base Transformer \cite{vaswani_attention_2017} parameters to be the most suitable for our problem.

\begin{figure}[H]
  \centering
  \includegraphics[keepaspectratio,scale=0.8]{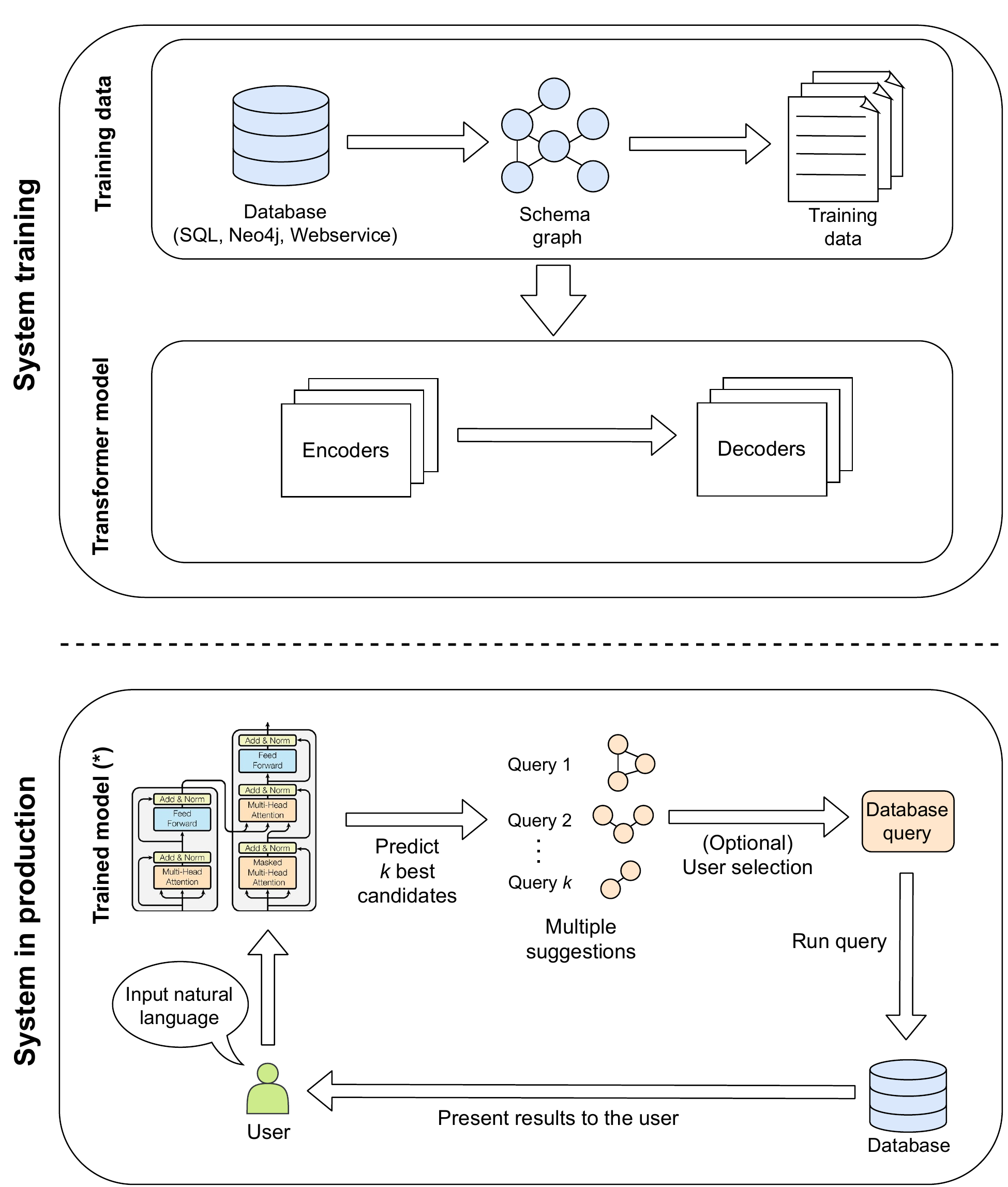}
  \caption{Graphical summary of the overall workflow for Polyglotter. (*) Adapted from Figure 1 of \cite{vaswani_attention_2017}}
  \label{fig:Figure2}
\end{figure}

Figure \ref{fig:Figure2} provides a graphical summary of the overall workflow for Polyglotter, using the Transformer as the translation model. First, training data was generated once from the target database using the random query generation procedure described in Section 2. Then, the model was trained with early stopping with the same procedure as for the original Transformer implementation, including the model’s hyperparameters.

The random walk procedure creates a bias against longer queries, yet these are of interest. Therefore we normalised the training set by the number of classes, setting a limit of five classes for HumanMine as this is greater than queries typically run on the database. Thus, for instance, the data set of size 100,000 for HumanMine was made up of five sets of 20,000 queries, with each set having a distinct number of classes from one to five. The class size cap was set according to the complexity of the database schema, at five for HumanMine (170 classes), four for MySQL (8 classes) and three for Neo4J (5 classes).

The datasets generated above were then divided into training, validation and test sets, comprising 60\%, 20\% and 20\% of the total observations, respectively. We evaluated the fraction of queries in the testing set, considering the classes and attributes, that were also in the training set for HumanMine (Supplementary Figure 2). Given that there are 170 classes, it is unsurprising that there is 21\% overlap when the dataset size is 1,000,000 and queries containing only one class ($nc$=1) are considered. This drops to 8.6\% for $nc$=2 and there is minimal overlap for $nc$=3,4,5 (2\%, 0.4\%, 0.07\% respectively). We note that for the simpler databases (MySQL: 8 classes; Neo4J: 5 classes) high test performance is achieved with datasets of only 5000 samples (see below) and at this dataset size, unsurprisingly we see overlaps in the range of 20-30\% for $nc$=1,2 and 6-7\% for $nc$=3.

The queries with the largest class count (5, 4, 3 for HumanMine, MySQL and Neo4j respectively) were kept in the testing set and never seen by the model during training, as a way of assessing the capability of the model to generalize to longer queries.

Once trained, and on receiving an input natural language query, the model is used to predict a set of $k$ query graphs as follows. The model directly predicts class-attribute pairs and class-attribute-constraint triples. In order to ensure the connectivity of these predictions, a minimum spanning tree of the predicted classes is obtained based on the database schema. Together with the predicted attributes and constraints this yields a query graph. Beam search was used to generate the top $k$ predictions for a given query \cite{Freitag_2017}. In our evaluation described in the following sections, we employ three values for $k$: $k$ = 1, 3, 5, corresponding to the top 1, top 3 and top 5 query graph suggestions, respectively. 

If given a choice in production, an end-user can choose which of these query graph predictions should be translated into a database-specific query that can be executed. We provide example implementations for transforming query graphs into equivalent database queries for the MySQL and Neo4j database engines, and also for HumanMine web services. It is also possible to translate the predicted query graph back into synthetic natural language. This can be used to help users select which of a number of alternative predictions is closest to what they want, and also to provide feedback to users who are trying to compose a query with any other query building tools.

\section{Evaluation of queries predicted from natural language}

We compared how the performance of Polyglotter varied with training dataset size in order to see whether there is evidence of saturation i.e. a point where increasing the number of instances available for training does not provide meaningful improvements in model accuracy. In order to assess the generalization performance we report the global accuracy by comparing the predicted query graphs against the known true query graph. We define the global accuracy over the testing set as the fraction of total testing set for which the predicted query graphs match exactly the ground truth for all the pairs and triples that comprise the query (i.e. predicting accurately all the attributes, classes and constraints).

\begin{figure}[H]
  \centering
  \includegraphics[keepaspectratio,scale=1]{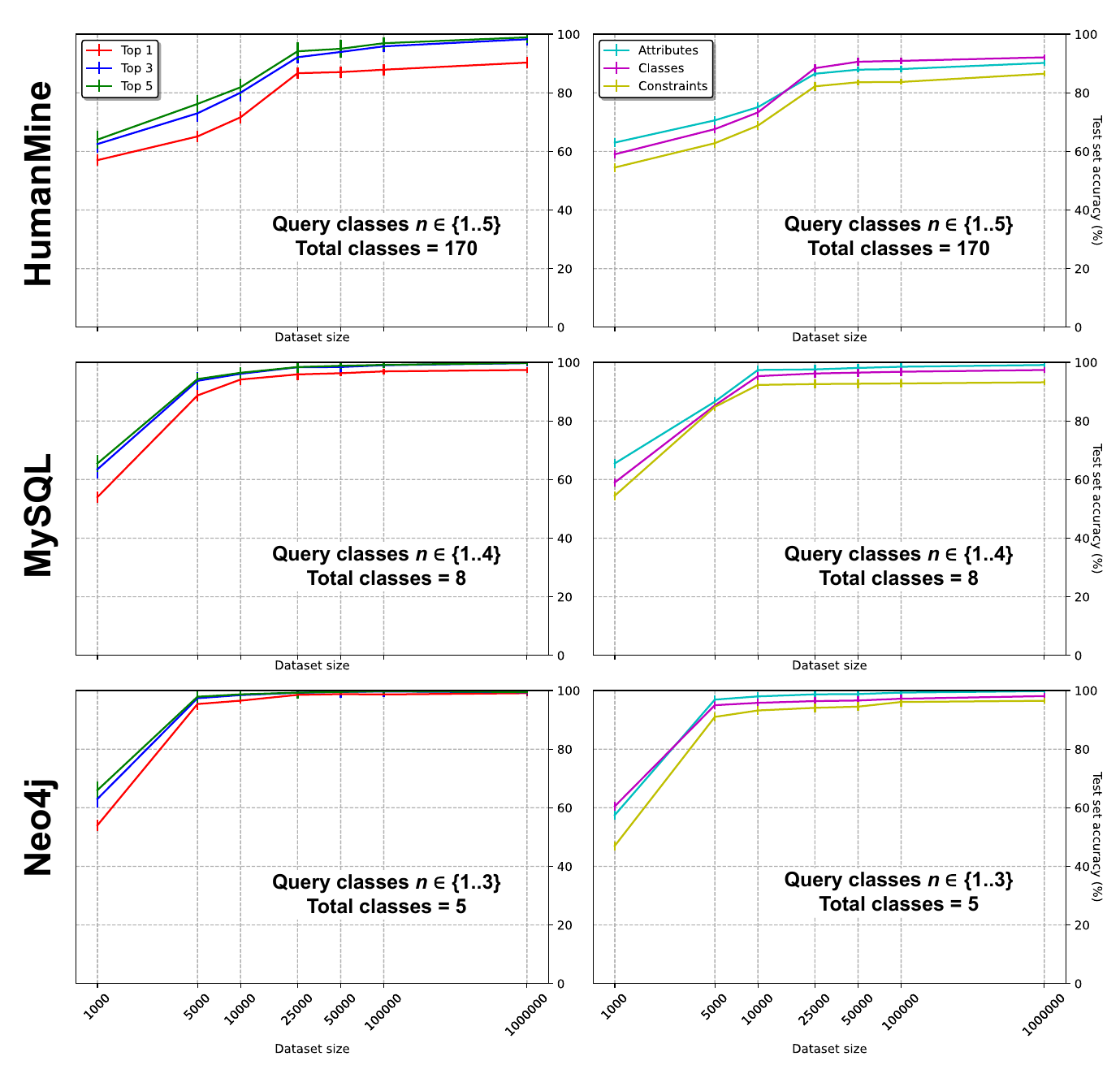}
  \caption{Test set performance as a function of training dataset size for HumanMine, MySQL and Neo4j (rows 1, 2 and 3 respectively). Left: overall test set performance allowing any of the top 1, 3 or 5 predictions from each model to match the corresponding test. Right: test set performance separately for each of the elements forming a query (attributes, classes and constraints) when using just the top prediction. The error bars show the standard deviation across ten independent datasets.}
  \label{fig:Figure3}
\end{figure}

Figure \ref{fig:Figure3} shows the test set performance of Polyglotter across different generated dataset sizes ($N$ = 1000 to 1,000,000) from the Neo4j, MySQL and HumanMine databases. As expected, we find that model accuracy increases with the size of the training corpus and that less complex databases require smaller training datasets to achieve a given level of performance. The impact of considering multiple predictions was examined by allowing any of the top 1, 3 or 5 predictions from each model to match the corresponding test, with overall performance increasing with the number of predictions considered.

In the case of HumanMine, substantial increases in accuracy are achieved until $N$ = 25000 (86.7\% test set accuracy with $k$ = 1), and more marginal but useful benefits are observed with larger datasets. For instance, increasing from $N$ = 25000 to $N$ = 1,000,000 (90.3\%) provides a gain of just 3.6\%, whereas from $N$ = 10000 (71.6\%) and $N$ = 25000 provides a gain of ~15\% in performance. For the other two databases we see similar trends, but in this case most of the performance has been achieved by $N$ = 5000 (MySQL 88.7\%; Neo4j 95.4\%). For these two databases, we note that from $N$ = 5000 the overlap of testing and training datasets is of slightly higher  degree compared to that of HumanMine at $N$ = 1,000,000. This is not surprising given that, for MySQL and Neo4j, the number of classes in the database schema is much lower (8 and 5 classes, respectively) than for HumanMine (170 classes). This means that the complexity of the set of queries is lower. For completeness we show the accuracies achieved up to $N$ = 1,000,000 for these databases (Figure 3).

As expected, increasing the number of predictions to three or five gives progressively better performance, with most of the improvement coming from considering three choices rather than just one. For HumanMine ($N$ = 1,000,000), relatively high performance is achieved with accuracies of 90.3\%, 98.3\% and 98.9\% for the top 1, top 3 and top 5 choices, respectively. Correspondingly, at $N$ = 5,000, the MySQL model reached 88.7\%, 93.7\% and 94.3\%, and the Neo4j model achieved 95.4\%, 97.3\% and 97.9\%. Across the three databases we note that the performance is not inversely related to the difficulty of the learning problem, represented by the complexity of the corresponding data model (i.e. The cardinality of the set of attributes and classes used in the database model). i.e. The performance on the most complex database (HumanMine) is intermediate to the two simpler ones with one choice, but better than either of them with 3 or 5 choices.

To assess the complexity of the query graph prediction problem, we evaluated separately how precise the models are at guessing each of the different components of the query graph i.e. the classes, attributes and constraints. Figure \ref{fig:Figure3} (right column) shows for each database the test set performance across the different generated datasets for each of these three components. From these experiments we note that predicting the constraints being expressed in the synthetic natural language queries was harder than predicting the classes and attributes. i.e.  HumanMine 88.5\% ($N$ = 1,000,000); MySQL 88.7\% and Neo4j 95.4\% for $N$ = 5,000 (Figure \ref{fig:Figure3}, second column - rows 1-3 respectively). Two possible explanations include, first, since the vocabulary that has to be predicted in the case of the constraints is infinite, it is more difficult for the model to achieve precise predictions. Second, in order to predict a constraint correctly, the class, attribute, constraint logic and constraint value must all be correct and this is harder than simple predicting class-attribute pairs. 

We also observed that predicting the attributes (HumanMine 90.2\% for $N$ = 1,000,000; MySQL 86.6\%; and Neo4j 96.9\% for $N$ = 5,000) was more difficult than predicting the classes (HumanMine 92.1\% for $N$ = 1,000,000; MySQL 87.3\%; Neo4J 97.4\% for $N$ = 5,000). This is probably because, with the random query generation parameters used, the number of attributes in a query is generally larger than the number of classes, posing a greater prediction challenge. We note that HumanMine has 103 unique attributes (1003 in total as some, such as “Name” or “Primary Identifier”, are used many times), while the MySQL and Neo4J examples have 59 and 42 attributes respectively, all of which are unique to the given example.

\begin{figure}[H]
  \centering
  \includegraphics[keepaspectratio,scale=0.5]{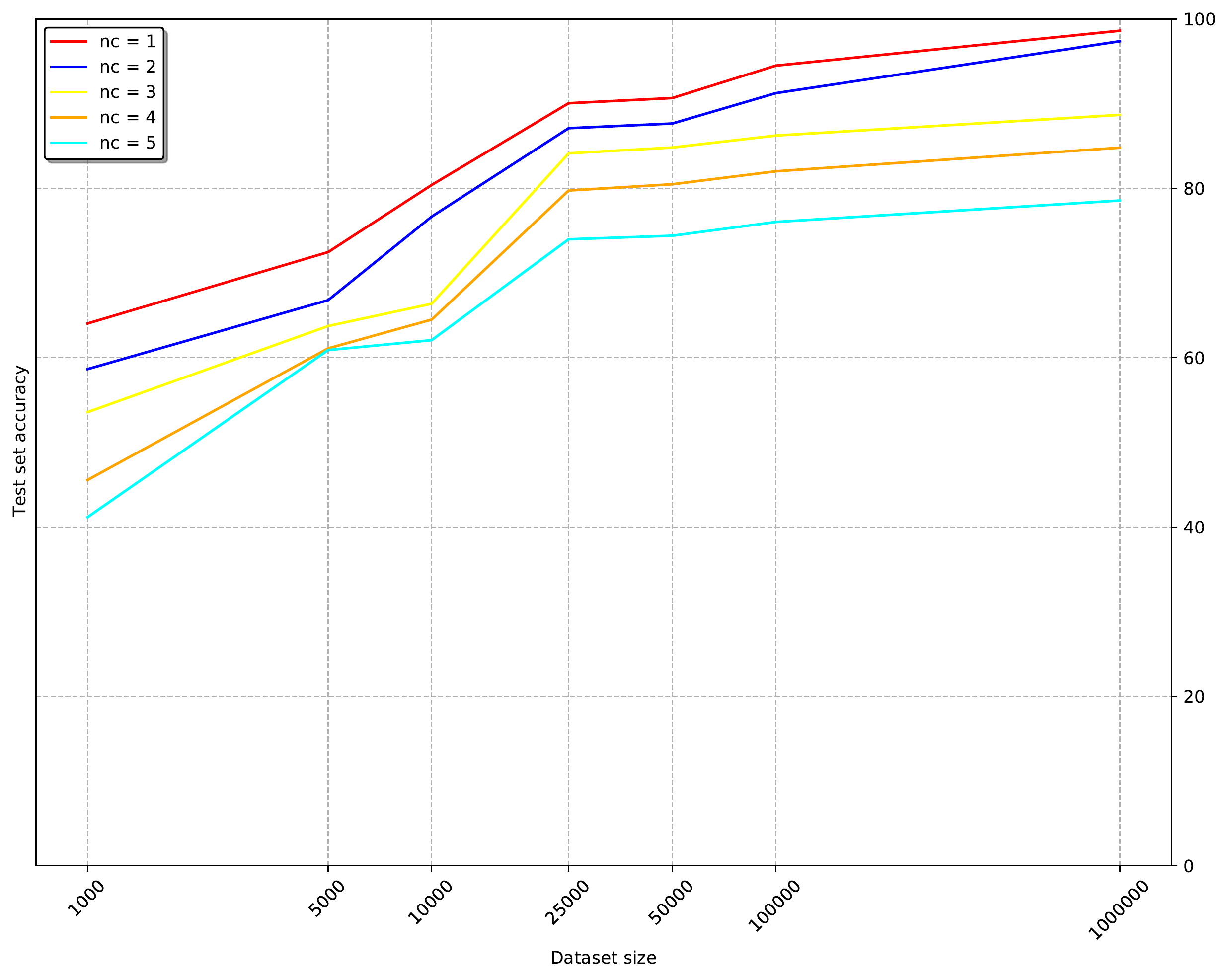}
  \caption{HumanMine performance as a function of number of classes in the query, as the number of training items varies, with $k$ = 1. Note that queries containing 5 classes were absent from the training set.}
  \label{fig:Figure4}
\end{figure}

The number of classes in a query is one proxy for query complexity. In order to understand how this affects performance in the most complex case examined, we compared the influence of training dataset size on the performance of queries containing different numbers of HumanMine classes (see Figure \ref{fig:Figure4}). As expected, queries with the smallest number of classes ($nc$ = 1 and $nc$ = 2) performed consistently better than the rest of the queries, achieving test set accuracies of 98.6\% and 97.4\% respectively, for $N$ = 1,000,000. Also, it’s worth noting that from $N$ = 50000, accuracies of over 90\% are already achieved for such cases. For queries covering 3, 4 and 5 classes, the accuracies achieved are noticeably lower: 88.7\%, 84.8\% and 78.6\%, respectively. Note that although queries containing five classes were included in the overall dataset, they were excluded from the training set. Thus this gives a measure of the ability of the model trained on equal weighting of classes 1-4 data to generalise to five classes.

In terms of training time, we note that, even for the most complex database (HumanMine) and the largest dataset size (N = 1,000,000), training took around seventeen hours on a single low-end GPU system (workstation with a NVIDIA GeForce RTX 2070 GPU; Supplementary Figure 3). Therefore, for database schemas of the complexity considered here, training does not require special resources.

\section{Conclusions and future work}

We have described a flexible framework, Polyglotter, for building natural language interfaces to multiple database types. In Polyglotter, we abstract the concepts of query and database schema in the form of graphs. We build a schema graph from any supported database engine, where the nodes are the original classes in the database, connected by edges depicting the class-class relationships, and with further edges to the attributes belonging to each class. User queries are abstracted into a labelled subgraph of the schema graph, containing attributes, classes and constraints. These query graphs can in turn be represented in the form of class-attribute pairs and class-attribute-constraint triples. During training, the Transformer model learns how to predict each of the pairs and triples necessary to reconstruct the query graph of the original synthetic natural language query. In this way Polyglotter  can translate the query graph abstraction to any supported database-specific query language. We provide support for two widely used engines (MySQL, Neo4j), and an example implementation for web service-based back-ends. The framework is easily extensible to further database engines.

Given that Polyglotter utilises a graph as an abstraction to represent queries, it can be used as a polyglot mediator, where a natural language query can be transformed into equivalent queries over multiple databases whether or not they use the same underlying engine. Thus Polyglotter has the potential to be an effective tool for cross-database querying and aggregation. In support of this we demonstrated that Polyglotter is capable of achieving good performance on three benchmark databases with different engines, with accuracies of 90.3\%, 88.7\% and 95.4\% in the HumanMine, MySQL and Neo4j databases, respectively and 98.3\%, 93.7\% and 97.4\% if the top three predictions are considered. These are assessed under conditions of limited overlap between the training and test datasets ($N$ = 1,000,000 for HumanMine, $N$ = 5,000 for the others). Thus, perhaps surprisingly,  performance is not closely related to the complexities of the underlying databases and is achieved without taking into account the typing of the individual attributes - doing so might yield modest improvements at the cost of a harder learning task.

As noted above on the real-world HumanMine database schema of 170 classes and over 1000 non-unique attributes, Polyglotter performance is 90.3\% when considering one prediction but increases to 98.28\% when considering the top three predictions. This demonstrates the value of giving users a choice of predictions before the query is executed, so that corrections can be made if needed, and exploits the inherently probabilistic nature of the Transformer model to provide these choices. Note that the predicted queries can be presented to the end user after translation back into synthetic natural language, and that this kind of natural language feedback can also be generated when the user is attempting to use conventional query building tools. Such feedback should help users who are learning to use query interfaces to determine whether they are moving in the intended direction.

The framework we describe does not come without drawbacks. Although we use a pre-trained FastText word embedding \cite{bojanowski2016enriching} to increase the chance of capturing semantic and lexical similarities across potential questions that have the same meaning, our current procedure to generate the training data from the database schema is based on a simple language model, where the main source of variability comes from permuting the different elements in the query. We try to alleviate this limitation a little by the use of synonyms in the question templates. Hence, there is an opportunity to improve overall performance and utility through improvements to the synthetic natural language generation procedure, and, for instance, generative adversarial network models might be an option worth exploring \cite{subramanian_adversarial_2017}. However we also note that humans are good at learning dialects and understanding broken natural language, and thus the quality of the generated natural language, whether for training or as an explanation for end-users, may not be such an important issue in practice. An advantage of the current simple model is that it should be relatively straightforward to add support for other languages and also the code is written in such a way that it would be easy to substitute an alternative approach to synthetic language generation.

The current approach is limited in the sense that the current random query generation process does not have the expressive power of the underlying database languages. Solving this completely would be a difficult task, but it should be relatively straightforward, for instance, to extend the query generator to include aggregation operators. Another extension to the system could be the capability to understand queries about the database structure itself. 

Notwithstanding the above limitations, we hope that the ability to train Polyglotter without the need for an annotated dataset will be a useful feature in practice, and one that will increase the opportunities for Polyglotter to provide natural language interfaces to a wider audience of database maintainers.

Finally, we believe that a separate study is needed to explore the effectiveness of Polyglotter as assessed by human users, and plan to recruit a small cohort of biologists to provide feedback on how effective Polyglotter is in the context of the most complex database studied, HumanMine. 

\section*{Data availability}

The datasets used in this study are available at \href{https://github.com/AdrianBZG/Polyglotter}{https://github.com/AdrianBZG/Polyglotter}.

\section*{Code availability}

The code developed for Polyglotter is publicly available at the Github repository \href{https://github.com/AdrianBZG/Polyglotter}{https://github.com/AdrianBZG/Polyglotter} under an Open Source Initiative (\href{https://opensource.org/licenses}{https://opensource.org/licenses}) approved licence, LGPL 2.1.

\bibliographystyle{unsrt}

\begin{thebibliography}{10}

\bibitem{affolter_comparative_2019}
Katrin Affolter, Kurt Stockinger, and Abraham Bernstein.
\newblock A comparative survey of recent natural language interfaces for
  databases.
\newblock {\em The VLDB Journal}, 28(5):793--819, October 2019.

\bibitem{dar_frameworks_2019}
Hafsa~Shareef Dar, M.~Ikramullah Lali, Moin~Ul Din, Khalid~Mahmood Malik, and
  Syed Ahmad~Chan Bukhari.
\newblock Frameworks for {Querying} {Databases} {Using} {Natural} {Language}:
  {A} {Literature} {Review}.
\newblock {\em arXiv:1909.01822 [cs]}, September 2019.
\newblock arXiv: 1909.01822.

\bibitem{reshma_review_2017}
E~U Reshma and P~C Remya.
\newblock A review of different approaches in natural language interfaces to
  databases.
\newblock In {\em 2017 {International} {Conference} on {Intelligent}
  {Sustainable} {Systems} ({ICISS})}, pages 801--804, Palladam, December 2017.
  IEEE.

\bibitem{blunschi_soda_2012}
Lukas Blunschi, Claudio Jossen, Donald Kossmann, Magdalini Mori, and Kurt
  Stockinger.
\newblock {SODA}: generating {SQL} for business users.
\newblock {\em Proceedings of the VLDB Endowment}, 5(10):932--943, June 2012.

\bibitem{shah_nlkbidb_2013}
Axita Shah, Jyoti Pareek, Hemal Patel, and Namrata Panchal.
\newblock {NLKBIDB} - {Natural} language and keyword based interface to
  database.
\newblock In {\em 2013 {International} {Conference} on {Advances} in
  {Computing}, {Communications} and {Informatics} ({ICACCI})}, pages
  1569--1576, Mysore, August 2013. IEEE.

\bibitem{franconi_panto_2007}
Chong Wang, Miao Xiong, Qi~Zhou, and Yong Yu.
\newblock {PANTO}: {A} {Portable} {Natural} {Language} {Interface} to
  {Ontologies}.
\newblock In Enrico Franconi, Michael Kifer, and Wolfgang May, editors, {\em
  The {Semantic} {Web}: {Research} and {Applications}}, volume 4519, pages
  473--487. Springer Berlin Heidelberg, Berlin, Heidelberg, 2007.
\newblock ISSN: 0302-9743, 1611-3349 Series Title: Lecture Notes in Computer
  Science.

\bibitem{sinha_connectionist_1994}
Chris Sinha, Ronan~G. Reilly, and Noel~E. Sharkey.
\newblock Connectionist {Approaches} to {Natural} {Language} {Processing}.
\newblock {\em The American Journal of Psychology}, 107(2):291, 1994.

\bibitem{deng_deep_2018}
Li~Deng and Yang Liu, editors.
\newblock {\em Deep {Learning} in {Natural} {Language} {Processing}}.
\newblock Springer Singapore, Singapore, 2018.

\bibitem{young_recent_2018}
Tom Young, Devamanyu Hazarika, Soujanya Poria, and Erik Cambria.
\newblock Recent {Trends} in {Deep} {Learning} {Based} {Natural} {Language}
  {Processing} [{Review} {Article}].
\newblock {\em IEEE Computational Intelligence Magazine}, 13(3):55--75, August
  2018.

\bibitem{huang_deep_2019}
Haiqin Yang, Linkai Luo, Lap~Pong Chueng, David Ling, and Francis Chin.
\newblock Deep {Learning} and {Its} {Applications} to {Natural} {Language}
  {Processing}.
\newblock In Kaizhu Huang, Amir Hussain, Qiu-Feng Wang, and Rui Zhang, editors,
  {\em Deep {Learning}: {Fundamentals}, {Theory} and {Applications}}, volume~2,
  pages 89--109. Springer International Publishing, Cham, 2019.
\newblock Series Title: Cognitive Computation Trends.

\bibitem{dichev_deep_2016}
Alexander Popov.
\newblock Deep {Learning} {Architecture} for {Part}-of-{Speech} {Tagging} with
  {Word} and {Suffix} {Embeddings}.
\newblock In Christo Dichev and Gennady Agre, editors, {\em Artificial
  {Intelligence}: {Methodology}, {Systems}, and {Applications}}, volume 9883,
  pages 68--77. Springer International Publishing, Cham, 2016.
\newblock Series Title: Lecture Notes in Computer Science.

\bibitem{habibi_deep_2017}
Maryam Habibi, Leon Weber, Mariana Neves, David~Luis Wiegandt, and Ulf Leser.
\newblock Deep learning with word embeddings improves biomedical named entity
  recognition.
\newblock {\em Bioinformatics}, 33(14):i37--i48, July 2017.

\bibitem{bahdanau_neural_2016}
Dzmitry Bahdanau, Kyunghyun Cho, and Yoshua Bengio.
\newblock Neural {Machine} {Translation} by {Jointly} {Learning} to {Align} and
  {Translate}.
\newblock {\em arXiv:1409.0473 [cs, stat]}, May 2016.
\newblock arXiv: 1409.0473.

\bibitem{dong_language_2016}
Li~Dong and Mirella Lapata.
\newblock Language to {Logical} {Form} with {Neural} {Attention}.
\newblock {\em arXiv:1601.01280 [cs]}, June 2016.
\newblock arXiv: 1601.01280.

\bibitem{sutskever_sequence_2014}
Ilya Sutskever, Oriol Vinyals, and Quoc~V. Le.
\newblock Sequence to {Sequence} {Learning} with {Neural} {Networks}.
\newblock {\em arXiv:1409.3215 [cs]}, December 2014.
\newblock arXiv: 1409.3215.

\bibitem{zhong_seq2sql_2017}
Victor Zhong, Caiming Xiong, and Richard Socher.
\newblock {Seq2SQL}: {Generating} {Structured} {Queries} from {Natural}
  {Language} using {Reinforcement} {Learning}.
\newblock {\em arXiv:1709.00103 [cs]}, November 2017.
\newblock arXiv: 1709.00103.

\bibitem{xu_sqlnet_2017}
Xiaojun Xu, Chang Liu, and Dawn Song.
\newblock {SQLNet}: {Generating} {Structured} {Queries} {From} {Natural}
  {Language} {Without} {Reinforcement} {Learning}.
\newblock {\em arXiv:1711.04436 [cs]}, November 2017.
\newblock arXiv: 1711.04436.

\bibitem{yin_neural_2016}
Pengcheng Yin, Zhengdong Lu, Hang Li, and kao Ben.
\newblock Neural {Enquirer}: {Learning} to {Query} {Tables} in {Natural}
  {Language}.
\newblock In {\em Proceedings of the {Workshop} on {Human}-{Computer}
  {Question} {Answering}}, pages 29--35, San Diego, California, 2016.
  Association for Computational Linguistics.

\bibitem{weir_dbpal_2019}
Nathaniel Weir, Andrew Crotty, Alex Galakatos, Amir Ilkhechi, Shekar Ramaswamy,
  Rohin Bhushan, Ugur Cetintemel, Prasetya Utama, Nadja Geisler, Benjamin
  Hättasch, Steffen Eger, and Carsten Binnig.
\newblock {DBPal}: {Weak} {Supervision} for {Learning} a {Natural} {Language}
  {Interface} to {Databases}.
\newblock {\em arXiv:1909.06182 [cs]}, September 2019.
\newblock arXiv: 1909.06182.

\bibitem{kalderimis_intermine_2014}
Alex Kalderimis, Rachel Lyne, Daniela Butano, Sergio Contrino, Mike Lyne,
  Joshua Heimbach, Fengyuan Hu, Richard Smith, Radek Štěpán, Julie Sullivan,
  and Gos Micklem.
\newblock {InterMine}: extensive web services for modern biology.
\newblock 42:W468--W472.
\newblock \_eprint:
  https://academic.oup.com/nar/article-pdf/42/W1/W468/17422978/gku301.pdf.

\bibitem{klein-etal-2017-opennmt}
Guillaume Klein, Yoon Kim, Yuntian Deng, Jean Senellart, and Alexander Rush.
\newblock {O}pen{NMT}: Open-source toolkit for neural machine translation.
\newblock In {\em Proceedings of {ACL} 2017, System Demonstrations}, pages
  67--72, Vancouver, Canada, July 2017. Association for Computational
  Linguistics.

\bibitem{klein-etal-2018-opennmt}
Guillaume Klein, Yoon Kim, Yuntian Deng, Vincent Nguyen, Jean Senellart, and
  Alexander Rush.
\newblock {O}pen{NMT}: Neural machine translation toolkit.
\newblock In {\em Proceedings of the 13th Conference of the Association for
  Machine Translation in the {A}mericas (Volume 1: Research Track)}, pages
  177--184, Boston, MA, March 2018. Association for Machine Translation in the
  Americas.

\bibitem{klein-etal-2020-opennmt}
Guillaume Klein, Fran{\c{c}}ois Hernandez, Vincent Nguyen, and Jean Senellart.
\newblock The {O}pen{NMT} neural machine translation toolkit: 2020 edition.
\newblock In {\em Proceedings of the 14th Conference of the Association for
  Machine Translation in the Americas (Volume 1: Research Track)}, pages
  102--109, Virtual, October 2020. Association for Machine Translation in the
  Americas.

\bibitem{vaswani_attention_2017}
Ashish Vaswani, Noam Shazeer, Niki Parmar, Jakob Uszkoreit, Llion Jones,
  Aidan~N. Gomez, Lukasz Kaiser, and Illia Polosukhin.
\newblock Attention {Is} {All} {You} {Need}.
\newblock {\em arXiv:1706.03762 [cs]}, December 2017.
\newblock arXiv: 1706.03762.

\bibitem{Freitag_2017}
Markus Freitag and Yaser Al-Onaizan.
\newblock Beam search strategies for neural machine translation.
\newblock {\em Proceedings of the First Workshop on Neural Machine
  Translation}, 2017.

\bibitem{bojanowski2016enriching}
Bojanowski, Piotr and Grave, Edouard and Joulin, Armand and Mikolov, Tomas.
\newblock Enriching Word Vectors with Subword Information.
\newblock arXiv preprint arXiv:1607.04606.

\bibitem{subramanian_adversarial_2017}
Sandeep Subramanian, Sai Rajeswar, Francis Dutil, Chris Pal, and Aaron
  Courville.
\newblock Adversarial {Generation} of {Natural} {Language}.
\newblock In {\em Proceedings of the 2nd {Workshop} on {Representation}
  {Learning} for {NLP}}, pages 241--251, Vancouver, Canada, 2017. Association
  for Computational Linguistics.

\end{thebibliography}

\section*{Acknowledgements}

This work was supported by the Wellcome Trust [208381]; Innovate UK [KTP011266].

\section*{Author Contributions}

A.B. and G.M. designed the study. N.G. developed the initial version of the random query generator. A.B. further developed the random query generator, developed all other software including the training and prediction pipelines. A.B. and G.M. devised the experiments. A.B. performed the experiments. A.B. and G.M. interpreted the results and wrote the manuscript. G.M. conceived and supervised the project.

\section*{Competing Interests statement}

None declared.

\newpage

\section*{Supplementary Material}

\beginsupplement

\begin{table}[h]
\centering
\begin{tabular}{|l|} 
 \hline
show assignedby in hpoevidence \\
\hline
give confidence, description, name having primaryidentifier >= @value, type is @value in interactiondetail, interactionexperiment \\
\hline
from publication, give abstracttext, pages, issue with title lower than @value \\
\hline
what is primaryidentifier having secondaryidentifier >= @value from cds \\ [1ex] 
 \hline
\end{tabular}
\caption{Examples of synthetic English queries generated during the random generation process.}
\label{table:1}
\end{table}

\newpage
\begin{figure}[h]
  \centering
  \includegraphics[scale=0.8]{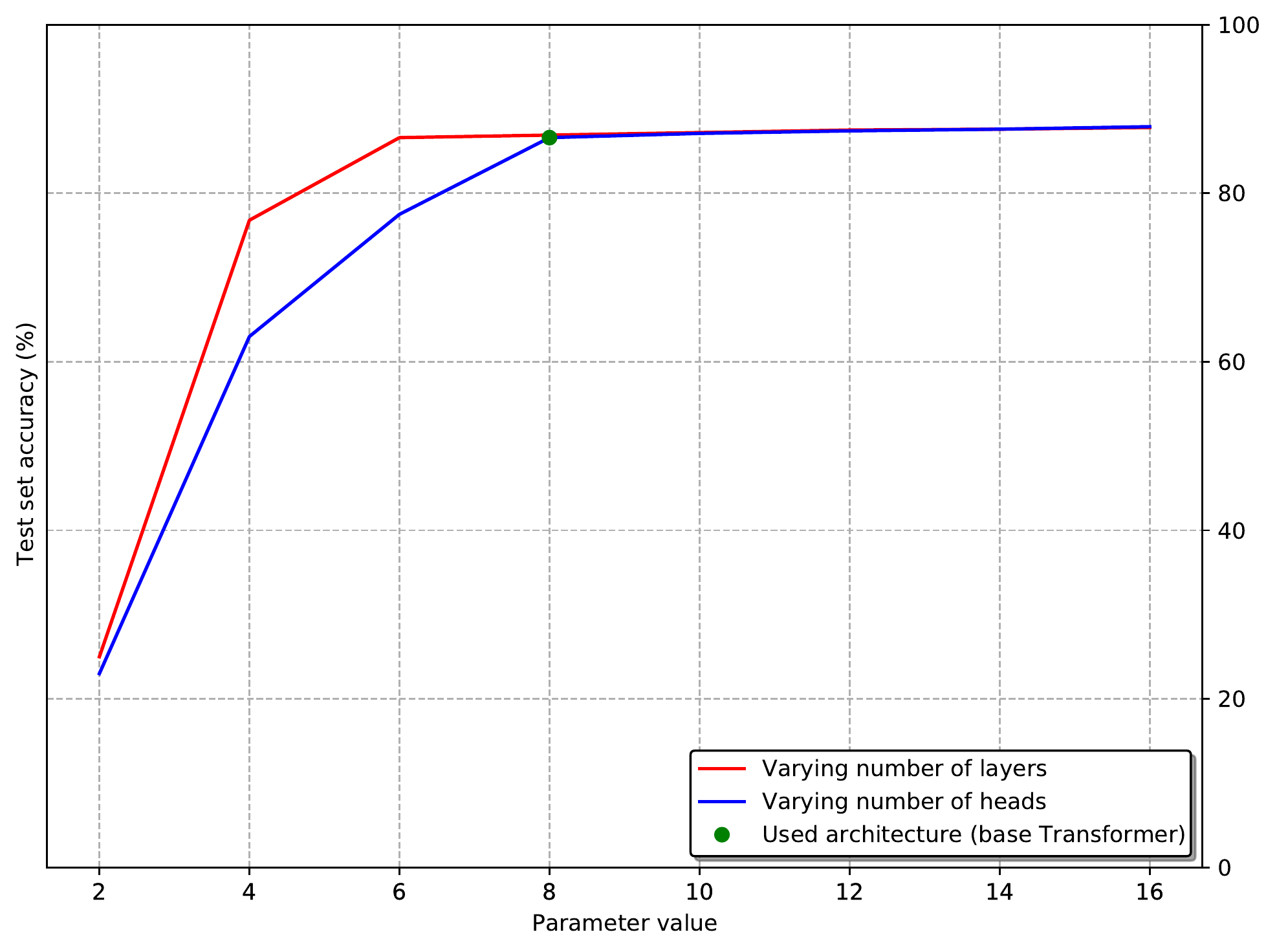}
  \caption{Ablation study for the Transformer model used in this paper, for HumanMine with N = 25000. Red line: test set performance for a varying number of network layers keeping the number of heads the same as the base Transformer (h = 8). Blue line: test set performance for a varying number of heads keeping the number of network layers the same as the base Transformer (L = 6). Green mark: Base transformer model, which was the selected architecture in this work.}
  \label{fig:Sup1}
\end{figure}

\newpage
\begin{figure}[h]
  \centering
  \includegraphics[scale=1]{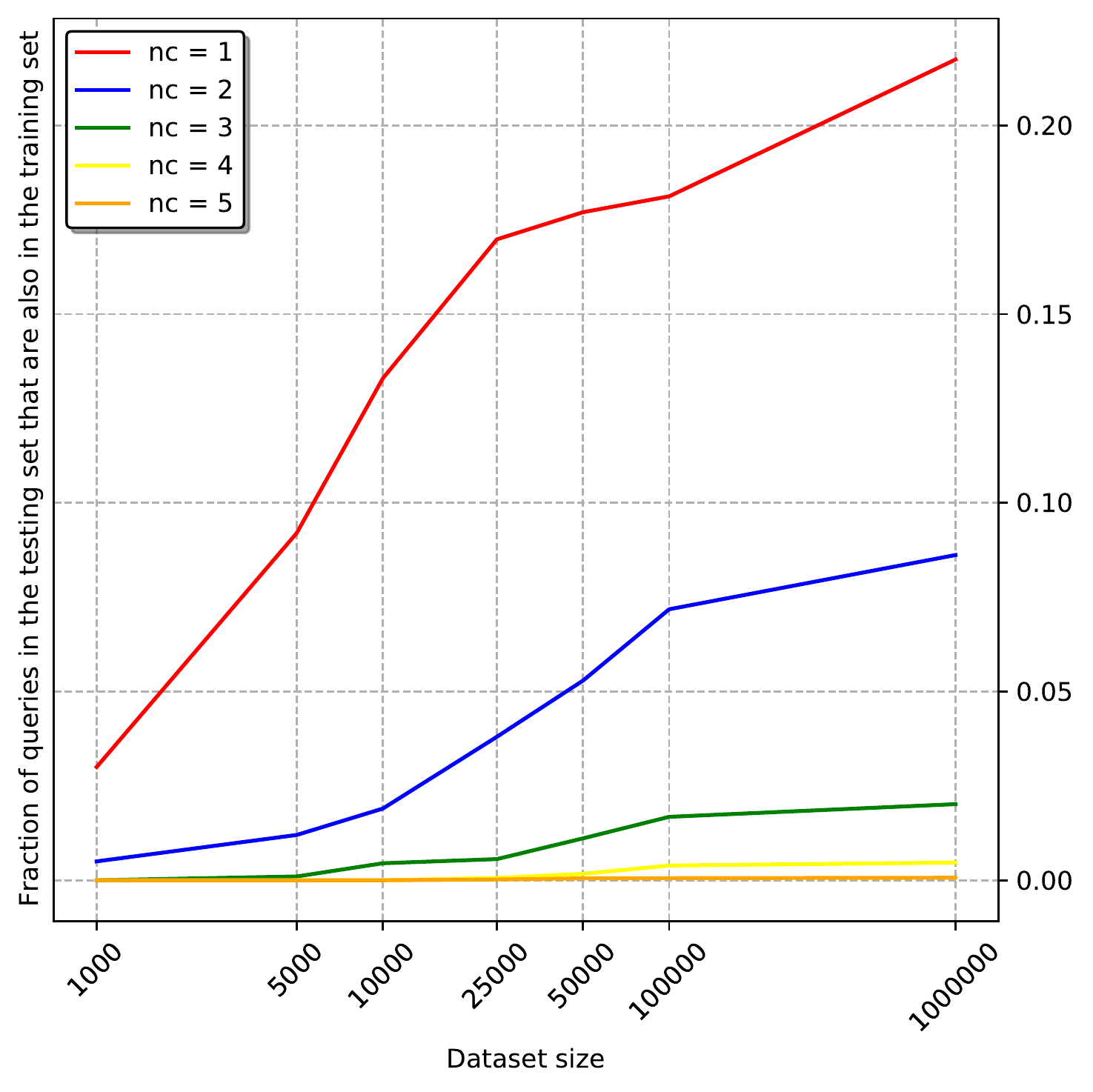}
  \caption{Overlap analysis: the HumanMine dataset was composed of equal numbers of queries covering 1-5 classes. We determined the degree of overlap between training and test sets by assessing the fraction of queries in the testing set that contained the exact same classes and attributes as a query or queries in the training set. This will be an overestimate of the true overlap as it does not take into account any constraints or constraint logic. Note that in fact nc = 5 was not used for training but the comparison is included here for completeness.}
  \label{fig:Sup2}
\end{figure}

\begin{figure}[h]
  \centering
  \includegraphics[scale=0.8]{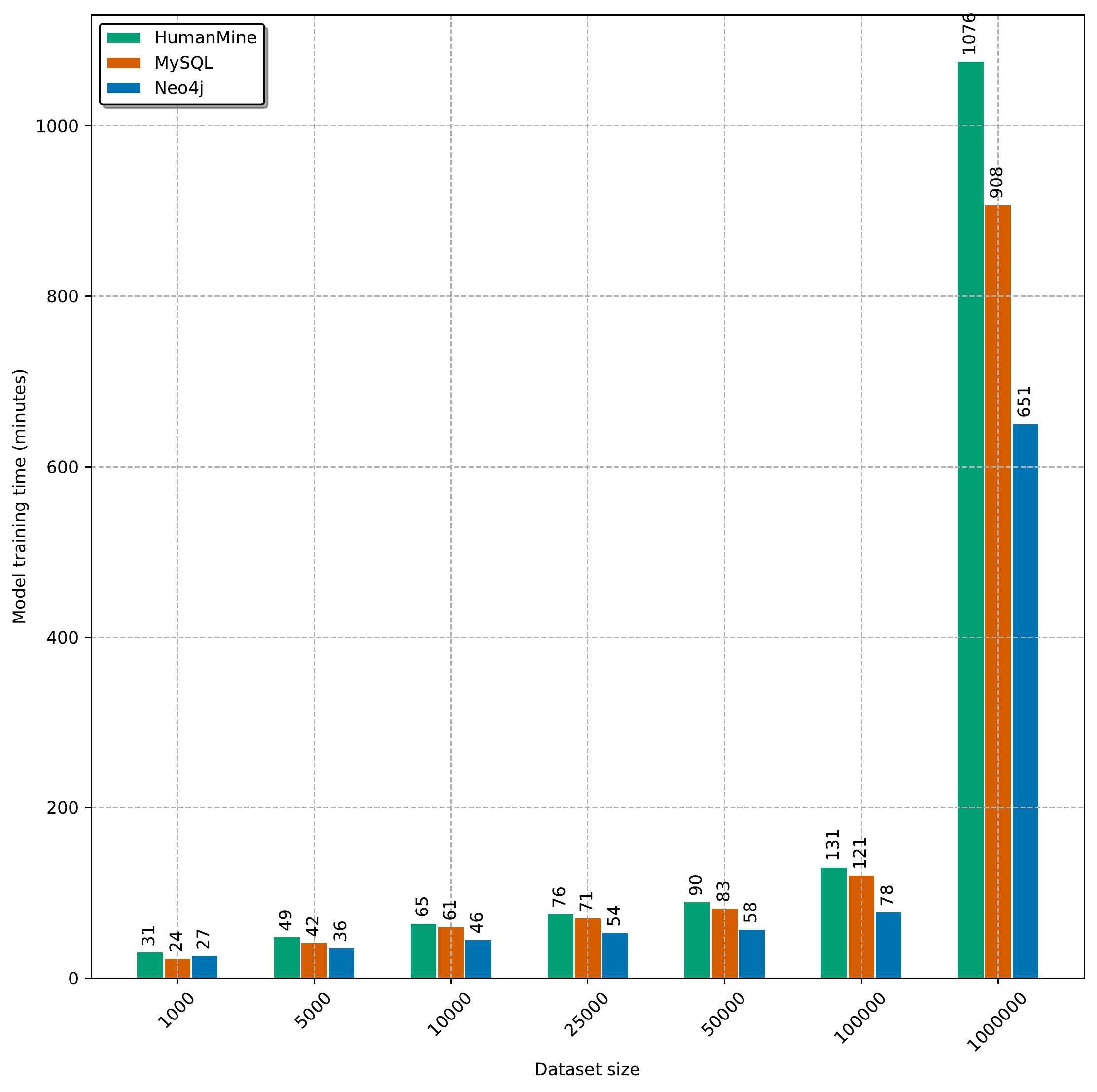}
  \caption{Model training times (in minutes) for increasing training dataset sizes across the three database systems tested. Training was carried out on a workstation with a single GPU (NVIDIA GeForce RTX 2070).}
  \label{fig:Sup3}
\end{figure}

\end{document}